\documentclass[sigconf]{acmart}
\AtBeginDocument{%
  \providecommand\BibTeX{{%
    \normalfont B\kern-0.5em{\scshape i\kern-0.25em b}\kern-0.8em\TeX}}}

\usepackage{amsmath}

\usepackage{amssymb}

\usepackage{multirow}

\renewcommand\footnotetextcopyrightpermission[1]{} 

\begin{document}

\title{Neural Network Quantization for Efficient Inference: \\ A Survey}

\author{Olivia Weng}
\affiliation{%
  \department{Dept. of Computer Science and Engineering}
  \institution{University of California, San Diego}
  \country{}
}
\email{oweng@ucsd.edu}

\renewcommand{\shortauthors}{Olivia Weng}

\newcommand{\blue}[1]{\textcolor{blue}{#1}}
\newcommand{\note}[1]{\blue{[#1]}}
\newcommand{\nts}[1]{\note{NTS: #1}}

\begin{abstract}
    As neural networks have become more powerful, there has been a rising desire to deploy them in the real world; however, the power and accuracy of neural networks is largely due to their depth and complexity, making them difficult to deploy, especially in resource-constrained devices. Neural network quantization has recently arisen to meet this demand of reducing the size and complexity of neural networks by reducing the precision of a network. With smaller and simpler networks, it becomes possible to run neural networks within the constraints of their target hardware. This paper surveys the many neural network quantization techniques that have been developed in the last decade. Based on this survey and comparison of neural network quantization techniques, we propose future directions of research in the area. 
\end{abstract}


\maketitle
\pagestyle{plain} 

\section{Introduction}
Neural Networks (NNs) are powerful tools for completing high-accuracy pattern recognition tasks; however, to achieve high levels of accuracy, NNs are often over-parameterized \cite{denil_predicting_2014}, growing to such size and depth that make it prohibitively expensive to deploy in resource-constrained environments, e.g., at the edge. Enabling neural network inference in resource-constrained settings is important so that NNs can solve problems like speech recognition, autonomous driving, and image classification in IoT devices, vehicles, and more. To realize this, neural network inference must achieve 1) real-time latency, 2) low energy consumption, and 3) high accuracy \cite{gholami_survey_2021}. 

It is difficult to achieve these three goals because NNs are designed in software to reach high accuracy by having tens to hundreds of millions of parameters. Moreover, when training NNs (which usually occurs on GPUs), these parameters are represented using 32-bit floating point numbers because of their high precision. When it comes time to deploy neural networks in hardware, however, we are left with massive NNs that are up to hundreds of megabytes ---at times, even several gigabytes---in size and require billions of floating point operations. The two main issues with deploying NNs are that they:
\begin{itemize}
    \item are too big for hardware and
    \item use expensive data types.
\end{itemize}

NNs are oftentimes too big for hardware because the target edge devices do not have enough memory to store and run the model. NNs often use 32-bit floating point data types to represent their parameters as well as all of the computations involved with inference, meaning they require expensive floating point units to run. These large sizes and intensive computation requirements present substantial obstacles to achieving fast and efficient inference.

Therefore, the size and complexity of NNs are major problems that inhibit real-time, efficient inference. To address these issues, several research avenues have emerged, such as
\begin{enumerate}
    \item designing/searching for novel small NNs,
    \item knowledge distillation, 
    \item pruning/sparsification, and
    \item quantization.
\end{enumerate}
All of these approaches have a common goal of reducing the size and/or complexity of NNs while maintaining high accuracy.

This paper focuses on NN quantization and surveys the numerous techniques that been developed over the years to address the associated challenges that have arisen in the space. Quantization is defined as reducing the precision used to represent neural network parameters, usually from $n$ bits to $m$ bits, where $n > m$. There is unique opportunity to achieving efficient inference with quantization because hardware such as Field-Programmable Gate Arrays (FPGAs) and Application-Specific Integrated Circuits (ASICs) can be configured to use any kind of numerical representation such as floating point, fixed point, integer, and even custom data types.
Thus, the goal with NN quantization involves not only reducing the number of bits but also converting the original numerical representation to a less precise data type that is cheaper to implement in hardware. Doing so makes it is possible to achieve efficient, real-time inference. For example, a common case in NN quantization involves starting with 32-bit floating point representation (since that is generally what is used to train NNs on GPUs) and converting it to 8-bit integer representation. This is desirable because integer arithmetic is less complex than floating point arithmetic and thus faster to compute. With fewer bits to compute and a cheaper numerical representation, executing an 8-bit integer NN significantly lowers latency, energy consumption, and resource utilization.

The primary challenge with quantization, however, is maintaining the NN's high accuracy post quantization. When reducing the precision of the network, the NN is effectively losing information it learned during training. This presents the main trade-off involved with quantization: \textit{precision vs. accuracy}. A reasonable expectation is that we must trade accuracy for lower precision to achieve smaller models that can fit in hardware; however, as is often the case with NNs, there are no hard and fast rules. 

There exists a wrinkle in the problem of NN quantization: NNs are often over-parameterized and can thus afford to lose precision with minimal to no loss in accuracy \cite{gholami_survey_2021}. Since one of the primary overarching goals in NN quantization is maintaining accuracy, we are not particularly wedded to quantizing weights such that their quantized values are as close to their floating point counterparts as permitted by the quantization scheme. We want to quantize the weights in a way that maintains the network's overall classification accuracy. In fact, trying to minimize the distance between the floating point and quantized weight representations does not directly translate to maintaining classification accuracy because of the over-parameterization of neural networks \cite{nagel_up_2020, gupta_deep_2015}. Therefore, the over-parameterization wrinkle presents opportunities to reduce precision in clever ways without any cost to accuracy---at times, quantizing a NN even improves its accuracy. 

NN quantization emerged as a field of study in the 1990s during a resurgence of neural network research \cite{hammerstrom_vlsi_1990, holt_finite_1993, hoehfeld_learning_1992}. 
At the time, one of the main inhibitors to neural network adoption in the real world was that training was too slow on the machines available then. As such, the primary motivation behind NN quantization then was to reduce the bit widths/precision as a way of speeding up training times. In 2012, AlexNet, a convolutional neural network (CNN), won the ImageNet Large Scale Visual Recognition Challenge, demonstrating that deep and complex networks could be trained efficiently on GPUs to achieve high accuracy. This popularized using GPUs to train NNs, overcoming the training bottleneck of the past and leading to an exponential increase in NN research. With so many new kinds of NNs being developed, companies and scientists have been itching to deploy them in hardware, such as edge devices, leading to a renewal in NN quantization efforts in the last decade. 

This paper surveys the NN quantization research that has surfaced in the last decade. It does not capture the full spectrum of NN quantization, for the field is massive. Therefore, we survey some of the most popular quantization techniques, namely integer/fixed point, binary, ternary, and mixed precision quantization. In our survey, we make the assumption that the reader has a basic understanding of NNs and their components, e.g., that NNs are made up of a series of weight and activation layers and trained with forwards and backward passes and weight updates via Stochastic Gradient Descent.

NNs can be quantized to make either training or inference more efficient (or both). In addition to quantizing a NN's parameters (namely its weights), quantized NN training quantizes gradients, whereas quantized NN inference quantizes activations. In this paper, since we are interested in the implications of quantizing NN for efficient hardware deployment, we limit our scope to inference, i.e., quantizing weights and activations, and only touch on training when it leads to hardware efficiency at inference time. 

\textbf{Paper Layout}: Section \ref{sec:background} gives an overview of how numbers are represented as data types on machines. Section \ref{sec:example} provides an example of quantization. Section \ref{sec:procedures} describes the two main procedures used to quantize a network, namely Quantization-Aware Training and Post-Training Quantization. Sections \ref{sec:int-fixed}, \ref{sec:binary-ternary}, \ref{sec:mixed} review work on four popular quantization schemes: integer/fixed point, binary, ternary, and mixed precision quantization. Section \ref{sec:future} concludes the paper with possible future avenues of research in NN quantization.

\section{Representing Numbers on Machines}\label{sec:background}
In this section, we review how data types are used to represent data, and more specifically real numbers, on machines. We detail how floating point and fixed point data types work because a common quantization scenario involves quantizing floating point to integer/fixed point. It is important to go into detail on the inner workings of floating point and fixed point to understand the costs and benefits of using them with respect to hardware.  

Since data on computers is represented in binary, the machine needs to know how to interpret the binary, i.e, as a character, string, integer, etc. This is the primary function of data types. Several standards have been developed to dictate how we should use binary code to represent different kinds of data. For instance, the ASCII standard defines a correspondence between natural numbers and individual characters, like letters, numbers, punctuation, etc. The same needs to be done for representing real numbers.

While integers have a straightforward correspondence to binary values, representing real numbers in binary is much more complex. There are several considerations to be made. For instance, how precise we want our real number representations to be, i.e., how many decimal places that should be accounted for. It is so complex that there is an IEEE standard for how real numbers should be represented as well as how arithmetic on this real number representation should work---IEEE Standard 754, also known as IEEE floating point. Since there are infinitely many real numbers and only so many bits that can be allocated for representing each number on machines, we can actually view representing real numbers on computers as a quantization problem itself because we are reducing the precision of the reals \cite{gholami_survey_2021}.

There are two main ways of representing real numbers: floating point and fixed point. 

\subsection{Floating Point}
The idea behind floating point is to represent numbers in scientific notation, $n \times 2^m$, wherein we need only keep track of $n$ and $m$ in our representation \cite{bryant_computer_2011}. IEEE floating point defines a floating point number $n$ as

\begin{align}
    n &= (-1)^s \times m \times 2^E 
\end{align}
where
\begin{itemize}
    \item $s$ is the \emph{sign} bit, 0 for positive and 1 for negative,
    \item $m$ is the \emph{mantissa}, also called the significand, which defines the precision, and
    \item $E$ is the \emph{exponent}, which determines the range.
\end{itemize}

\begin{figure}[t]
    \centering
    \includegraphics[width=\linewidth]{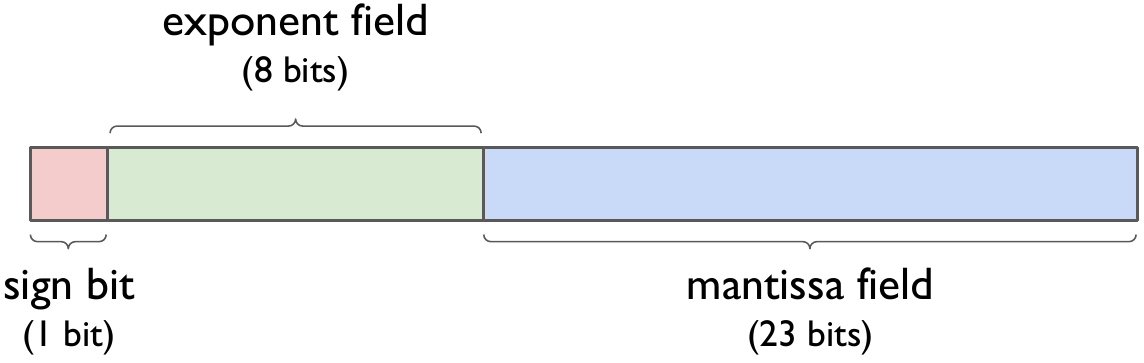}
    \caption{IEEE single-precision floating point bit fields. The binary of a floating point value is broken up, from least significant to most significant bit, into 23 bits for the mantissa field, 8 bits for the exponent field, and 1 bit for the sign bit.}
    \label{fig:float-bit-fields}
\end{figure}

A floating point number encodes the sign bit, mantissa, and exponent in its binary representation. In single precision floating point---the most common precision used for NN training on GPUs---, 32 total bits are given, in which 8 bits are allocated for the exponent, 23 bits are allocated for the mantissa, and 1 bit is allocated for the sign bit, as seen in Figure \ref{fig:float-bit-fields}. Depending on the value in the exponent field, there are three ways to interpret the values encoded in the exponent and mantissa bit fields (the sign bit is always interpreted as 0 or 1): 1) normalized values, 2) denormalized values, and 3) special values.

\subsubsection{Normalized Values} When the exponent field $E$ is neither 0 (all zeroes) nor 255 (all ones), we are working with normalized values. In this case, the bit fields are interpreted as follows:
\begin{itemize}
    \item $E = x - Bias$, where $x$ is the unsigned integer actually stored in the exponent bit field, and $Bias$ is $2^8 - 1 = 255$ (since we are given 8 exponent bits). Thus, $-254 \leq E \leq 255$. 
    \item $m = 1 + f$, where $f$ is the fractional binary value actually stored in the mantissa bit field. By fractional, we mean $0 \leq f < 1$. Essentially, $f$ is the fractional number to the right of the radix point. Thus, $1 \leq m < 2$. We implicitly add 1 to gain an extra free bit of precision. 
\end{itemize}

Based on these definitions, we compute normalized values $n$ from the bit fields as
\begin{align}\label{eq:norm}
    n &= (-1)^s \times (1 + f) \times 2^{x - 255}
\end{align}

As seen in Equation (\ref{eq:norm}), the floating point number is effectively scaled based on the value $x$ stored in the exponent field. This allows the radix point to ``float'' to various positions in the binary value as needed, providing floating point with a large range and thus more precision.

\subsubsection{Denormalized Values} When the exponent field $E$ is all zeroes, we are working with denormalized values. In this case, the bit fields are interpreted as follows:
\begin{itemize}
    \item $E = 1 - Bias$, where $Bias$ is still $2^8 - 1 = 255$, so $E = -254$. 
    \item $m = f$, where $f$ is the fractional binary value stored in the mantissa bit field, as in the normalized value case. 
\end{itemize}

Based on these settings, we compute denormalized values $m$ from the bit fields as 
\begin{align}\label{eq:denorm}
    n &= (-1)^s \times f \times 2 ^{-254}
\end{align}

\subsubsection{Special Values} When the exponent field $E$ is all ones, we represent special non-numerical values. For completeness, we include these special values. The value is determined by the bits residing in the mantissa bit field:
\begin{itemize}
    \item $m = 0$. Depending on the sign bit, this value is $-\infty$ or $+\infty$.
    \item $m \neq 0$. This is $NaN$, also known as Not a Number.
\end{itemize} 

Based on Equations (\ref{eq:norm}) and (\ref{eq:denorm}), we see that there is extra arithmetic on fractional binary values involved with floating point values. Even though much of this arithmetic has been optimized, there is still an overhead cost that needs to be paid when using floating point values; case in point, a floating point unit is required, making floats significantly more expensive to compute in hardware than say integers, which have no such overhead. Moreover, there are as many normalized values ($n \geq 1$) as there are denormalized values ($0 \leq n < 1$), meaning floating point numbers are non-uniformly spaced. There are many more values represented between 0 and 1 such that floating point has higher precision for values that lie in this range. This has many implications for quantization when quantizing from floating point to integer. For instance, if a network has many high-precision values between 0 and 1 and relies on this precision, this presents an extra challenge for quantizing to a less precise data type like integers.

\begin{figure}[t]
    \centering
    \includegraphics[width=\linewidth]{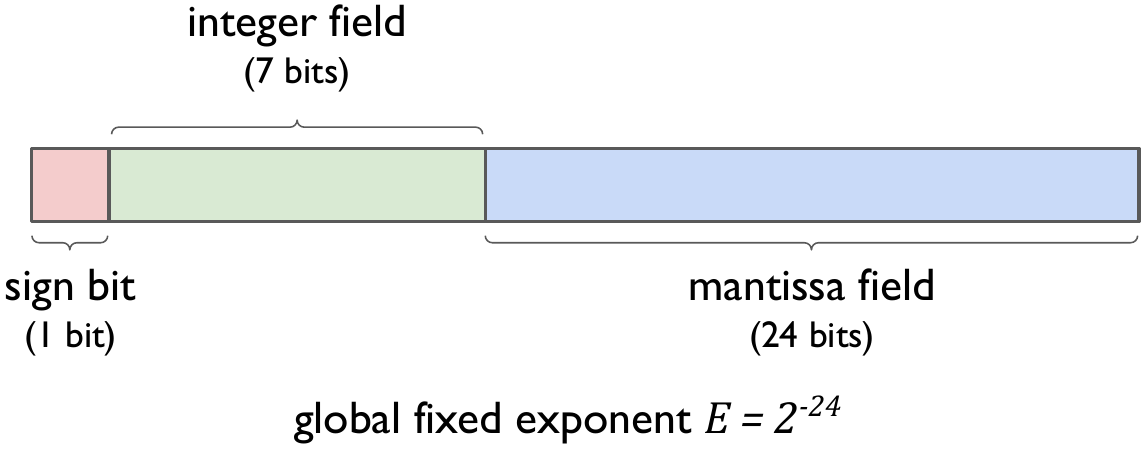}
    \caption{Example 32-bit fixed point format, where, from least significant to most significant bit, 24 bits are chosen for the mantissa, 7 bits are chosen for the integer, and 1 bit is for the sign bit. The value of the global fixed exponent depends on the number of bits allocated for the mantissa. In this case, there are 24 mantissa bits, implying that the global fixed exponent is $2^{-24}$.}
    \label{fig:fixed-bit-fields}
\end{figure}

\subsection{Fixed Point}
Fixed point is similar to floating point in that it is also encoded using a sign bit and a mantissa, but it uses a single global fixed exponent value that is shared across all fixed point values. Since the exponent is fixed, there is no need to store it in the binary, and the remaining bits are allocated as the integer field. Figure \ref{fig:fixed-bit-fields} depicts an example 32-bit fixed point value that is broken up into 24 mantissa bits, 7 integer bits, and 1 sign bit. The global exponent effectively places the radix point at a fixed position; hence, ``fixed point.'' The shared exponent is also referred to as a global scaling factor and is typically a power of 2, so that the scaling multiplication can be implemented using bit shifts \cite{courbariaux_training_2015}. Based on the fixed point definition in Figure \ref{fig:fixed-bit-fields}, given 24 mantissa bits, the global scaling factor is $2^{-24}$.  

Depending on the application, we might want more or less precision in our fixed point values. For instance, Vivado High-Level Synthesis defines a fixed point data type for its FPGA synthesis called \texttt{ap\_fixed$<T, I>$} that allows users to choose the fixed point format, where $T$ is the total number of bits allocated for the fixed point number and $I$ is how many integer bits are allocated \cite{vivado_user_guide}. Since the global exponent is fixed and is based on how many bits are allocated to the mantissa, there is a trade-off involved in choosing $T$ and $I$. Fewer integer bits means there is more bits left for the mantissa, implying high precision but a small range. More integer bits means there are fewer bits left for the mantissa, leading to low precision but a larger range. 

Fixed point is typically used in embedded systems that cannot afford to have a floating point unit but still need some precision in their computations \cite{courbariaux_training_2015}. Without a floating point unit, the fixed point bit fields are all interpreted as integers (as opposed to fractional binary values like the floating point mantissa is), which are much cheaper to compute on. To explain how this works, let us walk through an example. Based on the fixed point format depicted in Figure \ref{fig:fixed-bit-fields}, let us encode $\frac{1}{3}$. Since our value is a fraction, we do not need any integer bits, so this field is all zeroes. For the mantissa field, given 24 bits and a global exponent of $2^{-24}$, we store the integer 5,592,405 in the mantissa field because $5592405 \times 2^{-24} \approx \frac{1}{3}$. This can be derived for any real number fraction $r$ and $m$ mantissa bits, where the corresponding fixed point mantissa value is $\lfloor r \cdot 2^m \rfloor$ \cite{ericson_chapter_2005}. As such, to determine based on the sign bit, integer field, mantissa field, and global exponent the fixed point value $n$, we compute

\begin{align}
    n &= (-1)^s \times (I + m \times E)
\end{align}
where $s$ is the sign bit, $I$ is the integer encoded value, $m$ is the mantissa encoded value, and $E$ is the global exponent. Although this equation looks similar to the floating point equations, note that all of the values are integers and $E$ is typically a power of 2, so all the computation can be performed using integer arithmetic and bit shifts, which is significantly cheaper than floating point arithmetic.


\section{Quantization Example}\label{sec:example}
To quantize a value, we must follow a quantization function that systematically maps high precision values to low precision values. 

\subsection{Quantization}
A common quantization function $Q(r)$ maps a floating point value $r$ to an integer by way of

\begin{align}\label{eq:quant}
    Q(r) &= Int(r / S) - Z
\end{align}
where $S$ is a floating point scaling factor and $Z$ is an integer zero point \cite{gholami_survey_2021}. By zero point, we mean $Z$ is the integer value that represents 0 in our quantization scheme, which could be 0 or another value. In our scope of quantizing NN for efficient inference, $r$ is either a weight or activation. The $Int(\cdot)$ function assigns the scaled $r$ to an integer, typically via rounding. The rounding function can be as simple as round-to-nearest or something more complex, as seen in \cite{nagel_up_2020, hubara_accurate_2021}. In this case, $Q(r)$ is an example of \textit{uniform quantization} because all of our $r$ input values are scaled by the same value $S$, implying that the distance between quantized values is equally spaced and thus uniform. It is possible to define non-uniform distances between quantized values, which is known as \textit{non-uniform quantization}; however, this is out of scope. 

\subsubsection{Choosing a Scaling Factor}
Choosing a scaling factor in a way that minimizes accuracy loss is non-trivial. The scaling factor plays a large role in quantization because, as previously discussed, the scaling factor determines the distance between quantized values, i.e., the step size. The scaling factor is defined as

\begin{align}\label{eq:scaling}
    S &= \frac{\beta - \alpha}{2^b - 1}
\end{align}
where $\beta$ is the upper bound and $\alpha$ is the lower bound of the range of quantized values, and $b$ is the quantization bit width. $[\alpha, \beta]$ is also known as the \textit{clipping range}. 

Determining the clipping range is known as \textit{calibration}. When $\alpha = -\beta$, we are employing \textit{symmetric quantization}. Using symmetric quantization means the zero point $Z$ is 0, which is computationally less expensive at inference time, especially since the quantization function is now merely $Q(r) = Int(r / S$. While symmetric quantization is more inexpensive, it can occur at the cost of accuracy, in particular for NNs that have an imbalance in their weights or activations. This is apparent in NNs that use ReLU activations, a common activation function, that results in only non-negative activation values. In response, we can select a clipping range such that $\alpha \neq -\beta$ that better reflects the imbalance of our weights and/or activations. This is known as \textit{asymmetric quantization}. More information on calibrating clipping ranges can be found in \cite{gholami_survey_2021}.

\subsection{Dequantization}
To go from the quantized value to its original real value, we use a dequantization function, which does the reverse of the quantization function. Dequantization is defined as

\begin{align}\label{eq:dequant}
    \hat r &= S \cdot (Q(r) + Z)
\end{align}
Note that $\hat r$ is not guaranteed to equal the original value $r$ because $Q(r)$ employs a rounding function. The bias introduced by the rounding function introduces some error that is lost and cannot be recovered by the dequantization function.\footnote{Some quantization work has focused on accounting for this rounding error by introducing bias into the NN's parameters \cite{fan_training_2021, zhou_dorefa-net_2018}.}

\section{Quantization Procedures}\label{sec:procedures}
\begin{figure}[t]
    \centering
    \includegraphics[width=\linewidth]{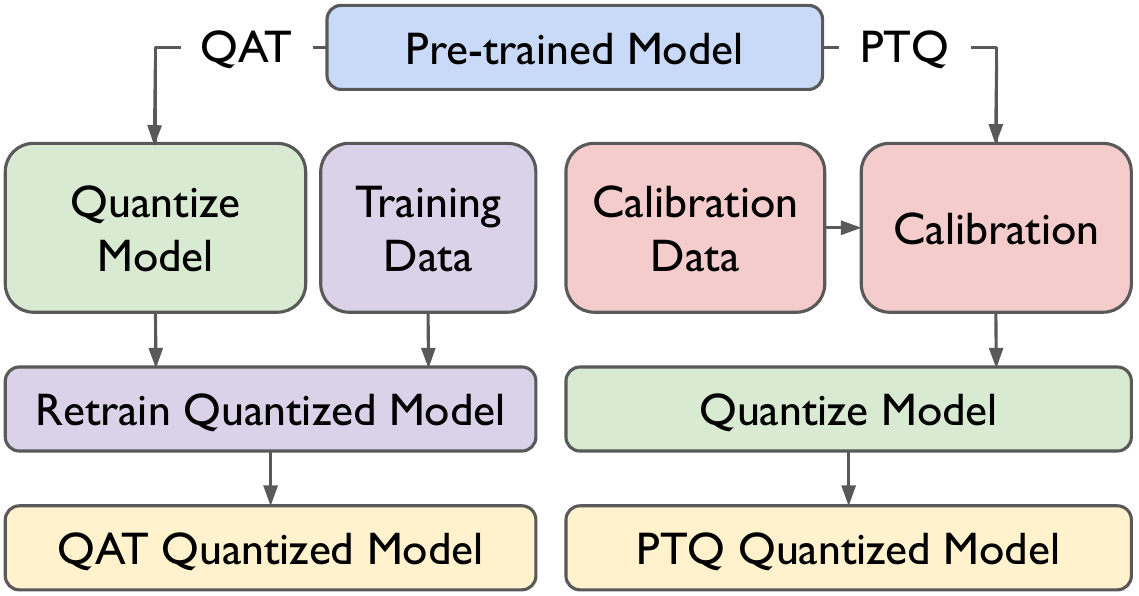}
    \caption{How to quantize a pre-trained model via either Quantization-Aware Training (QAT) or Post-Training Quantizatino (PTQ) \cite{gholami_survey_2021}. Calibration data can be either a subset of training data or a small set of unlabelled input data. Refer to Section \ref{sec:example} to review calibration.}
    \label{fig:quant-procedures}
\end{figure}

While it is possible to train a quantized NN from scratch, the majority of research has shown that starting with a pre-trained model to be more effective at minimizing accuracy loss when quantizing a NN. There exists two main methods of quantizing pre-trained NNs: quantization-aware training (QAT) and post-training quantization (PTQ). QAT involves quantizing a NN and then retraining it so that it has a chance to adjust and learn according to the newly quantized values. Sometimes a sufficient amount of training data is not readily available, so we use PTQ, which entails quantizing a NN without any extra training. Figure \ref{fig:quant-procedures} gives an overview of QAT and PTQ.

\begin{table*}[t!]
\caption{\label{tab:procedures} A qualitative comparison of the two main quantization procedures}
\centering
    \begin{tabular}{lccc}
        \hline
        \textbf{Quantization Procedure} & \textbf{Accuracy Loss} & \textbf{Quantization Time} & \textbf{Minimum Achievable Precision} \\
        \hline
        Quantization-Aware Training & Negligible & High & $\geq 1$ bit \cite{leibe_xnor-net_2016} \\
        \\
        Post-Training Quantization & Moderate & Low & $\geq 4$ bits \cite{nagel_up_2020} \\
        \hline
    \end{tabular}
\end{table*}

\subsection{Quantization-Aware Training}
Quantization-Aware Training involves retraining a model with quantized parameters so that it can learn and correct any quantization bias that often results from rounding errors. To perform QAT, we typically
\begin{enumerate}
    \item quantize the weights, 
    \item perform a forward training pass through the model using floating point inputs and activations,
    \item perform a backward pass through the quantized model using floating point gradients,
    \item update the weights using the floating point gradients, and proceed back to step (1).
\end{enumerate}
until the NN converges.

Regarding step (3), we note that the weights are quantized using a quantization function like the one defined in Equation (\ref{eq:quant}), which is non-differentiable. Thus, it is necessary to approximate the gradient. A popular heuristic used for this approximation is called the Straight-Through Estimator (STE) \cite{bengio_estimating_2013}. The STE approximates the non-differentiable parts of the quantization function using the identity function. While it is not clear why the STE works, it has been empirically shown to be effective, except for extreme bit widths (e.g., quantizing to 1-bit binary values).

In addition to correcting the quantization bias in weights, other quantization parameters can be learned during QAT, such as the clipping range $\alpha$ and $\beta$ as defined in Equation (\ref{eq:scaling}). For instance, Parameterized Clipping Activation (PACT) \cite{choi_pact_2018} learns the activation clipping ranges for each activation layer during QAT.

The main trade-off of the high accuracy afforded by QAT is the long training time, e.g., hundreds of epochs, and the associated computational retraining cost. Moreover, a sufficient amount of training is required to prevent ove-fitting. Nevertheless, this long training time is often worth it for models that will be deployed for long periods of time, wherein the hardware and energy efficiency gains more than make up for the retraining cost.

\subsection{Post-Training Quantization}
QAT not possible when the training data is too sensitive or unavailable at time of deployment. At times like these, Post-Training Quantization is an attractive option for fast quantization. Oftentimes, PTQ uses a small set of calibration data, such as unlabelled input data, to help with choosing the best quantization parameters, e.g., scaling factor. In the past, Post-Training Quantization penalized all quantization errors equally, which is less than ideal because some quantization errors contribute more to altering classification than others. As a result, researchers turned towards QAT to fix the error introduced by quantization by retraining the entire model, as previously discussed. Recently, however, people have been revisiting PTQ in an attempt to quantize in a smarter way when there is limited data available.

To apply PTQ to a NN, we typically 
\begin{enumerate}
    \item calibrate the quantization parameters based on any available calibration data, and then
    \item quantize the model.
\end{enumerate}

When there is a lack of training data and fast quantization is needed, PTQ is usually the best option. This is often at the cost of achievable precision in that at least 4 bits are required \cite{nahshan_loss_2020}. Even with 4 or more bits, PTQ tends to be less accurate than QAT, so it is the less popular option because with the same bit precision, QAT achieves higher accuracy. Table \ref{tab:procedures} summarizes these trade-offs.

\section{Integer/Fixed Point Quantization}\label{sec:int-fixed}
As described in Section \ref{sec:background}, fixed point values are computed using integers, so they cost relatively the same to execute as plain integers do, give or take a few extra bit shifts/multiplications to take care of the global scaling factor. To this end, we consolidate integer and fixed point quantization in the same section.

Since the goal if NN quantization is to reduce the precision of NN implementations, a natural first step is to see if lower bit width integer or fixed point values are sufficient for accurate classification because they are cheaper and faster in hardware than floating point is. Quantizing NNs from floating point values to integer was studied as a early as in the 1990s \cite{hammerstrom_vlsi_1990, hoehfeld_learning_1992, holt_finite_1993}. \cite{holt_finite_1993} showed that for the simpler networks at the time, it was possible to quantize to 8-bit integer with minimal accuracy loss. We have seen a continuation of that result in more recent work \cite{jacob_quantization_2018, zhuang_towards_2018}, with researchers now pushing the envelope and quantizing to as low as 2-bit integers \cite{fan_training_2021, zhuang_towards_2018, nahshan_loss_2020}. 

In the early 2010's, \cite{courbariaux_training_2015, gupta_deep_2015} explored training NNs with fixed point data types. While these works were focused on training, it is worth noting that \cite{courbariaux_training_2015} experimented with training fixed point as well as \textit{dynamic fixed point} \cite{williamson1991dynamically}. Dynamic fixed point attempts to meet floating point and fixed point in the middle. Recall from Section \ref{sec:background} that floating point encodes a unique exponent for each value, whereas fixed point shares a global exponent across all values. Dynamic fixed point employs several scaling factors that are shared among a group of values, and these scaling factors are dynamically updated as the statistics of the group changes. On Maxout networks \cite{goodfellow2013maxout}, the authors show that 10 bits were sufficient for minimal accuracy loss when using dynamic fixed point, compared with the 20 bits that were necessary with regular fixed point. Although they were able to reduce the bit precision to 10 bits, dynamic fixed point is quite expensive to implement in hardware because the shared exponents must all be updated every so often. This extra dynamic overhead is similar to that found in \textit{dynamic quantization} and does not lend itself well to hardware-efficient inference. Therefore, plain fixed point/integer is preferred.

\cite{jacob_quantization_2018} provides a QAT framework for integer NN quantization and is the basis for TensorFlow Lite \cite{tflite}, which targets NN inference on mobile and edge devices. During the normal training procedure in floating point, the quantization function is applied to the weights and activations to simulate quantized NN inference. Using this method, they quantize the weights to 8 bits, whereas the biases are quantized to 32 bits. For each array of weights, they learn the quantization scaling factor (Equation \ref{eq:scaling}) and zero point (Equation \ref{eq:quant}) during QAT. Based on these learned quantization parameters, they infer what the best quantization parameters are for the activation arrays. They also make use of ``folding'' the batch normalization (BN) parameters (when they are present) into the weights before applying the quantization function during training. The idea is to combine the BN parameters with the weight parameters, consolidating the two arrays into one parameter array to reduce multiplications. This way the input data is multiplied with the weights and the batch normalization parameters all at once rather than with each of the two arrays separately. To compute a folded batch normalization weight $w_{BN}$, do 
\begin{align}
    w_{BN} &= \frac{\gamma w}{\sqrt{\sigma^2_B + \epsilon}}
\end{align}
where $w$ is the original weight and $\gamma$, $\sigma_B$, and $\epsilon$ are the BN parameters. Doing so, in effect, ``pre-computes'' BN, reducing the number of multiplications at inference time. Therefore, the weights' quantization parameters reflect the effects of BN folding for more efficient inference.

Similar to \cite{jacob_quantization_2018}, \cite{gong_highly_2018} also fuses BN with the weights, taking the approach further by fusing BN with the biases, if they are used. A BN-fused bias $b_{BN}$ is computed as
\begin{align}
    b_{BN} &= (\frac{\gamma}{\sqrt{\sigma^2_B + \epsilon}})(b - \mu) + \beta
\end{align}
where $\mu$ and $\beta$ are additional BN parameters affecting the bias. A major difference with \cite{jacob_quantization_2018}, however, is that \cite{gong_highly_2018} is a PTQ technique. They use calibration data to compute the quantization scaling factors for the weights, activations, and biases. By passing the calibration data through the network, the authors record what the maximum absolute value $MAX$ of each array was. The scaling factor $S$ is then set to 
\begin{align}
    S &= \frac{MAX}{2^b - 1}
\end{align}
where $b$ is the desired bit precision. The quantize the weights and activations so that they fit in 8-bit integers and biases so that they fit in 32-bit integers. Since they assume ReLU activations, which are non-negative, they choose $b = 8$ for activations, $b = 7$ for weights, and $b = 31$ for biases---weights and biases to save one bit for the sign bit. They report accuracy losses of less than 1\%.

Another integer PTQ procedure called Loss-Aware Post-Training Quantization (LAPQ) \cite{nahshan_loss_2020} reduces the achievable bit width further to 4 bits. The authors identify a problem with prior PTQ methods: choosing the scaling factor for each tensor \textit{independently} when there is a quantization error dependency between layers. In essence, the quantization noise of one layer affects the next one and the one after, snow balling to the end of the network. This is particularly detrimental at lower bit widths (2-4 bits). Therefore, LAPQ optimizes the quantization parameters of a layer with the other layers in mind. The authors view quantization as a multivariate optimization problem. By combining traditional layer-by-layer PTQ with their multivariate optimization problem, LAPQ achieves better accuracy results for 4-bit weight and activation quantization than prior work at the time, though LAPQ's 8-bit weights and 4-bit activations are better at maintaining accuracy (around 2\% accuracy loss). Note that they do not quantize the first and last layers whatsoever, which is a major limitation, especially for target hardware that lack floating point units.

\cite{nagel_up_2020} is another PTQ method, dubbed AdaRound, that quantizes NNs to 4-bit integers with around 1\% accuracy loss. In this work, the authors show that for PTQ, rounding-to-nearest is not optimal and is a primary cause for the propagation of quantization/rounding errors through the network. As such, they pose rounding from floating point to integer as an optimization problem that, similar to LAPQ, takes into account the quantization errors from the preceding layers. This way AdaRound minimizes the accumulation of quantization errors that is especially apparent and incurs high accuracy loss in deeper networks. Thus, AdaRound adapts the rounding for each weight layer to the statistics of its input data with only a small amount of unlabelled calibration data; however, AdaRound only quantizes weights, when the remaining network parameters need to be quantized as well for hardware-efficient deployment.

Given the limitations of PTQ, in particular the lack of access to training data, Quant-Noise \cite{fan_training_2021} seeks to achieve 8-bit and 4-bit integer quantization via QAT; in fact, they incur minimal accuracy losses when training from a quantized model from scratch, i.e., they do not need to start with a pre-trained model. Their unique QAT procedure involves choosing a quantization method and applying it to a random \textit{subset} of weights during each forward pass of training so that that some gradients that are unbiased by the STE are allowed to update the weights. In effect, they are introducing the noise produced by the chosen quantization method randomly and incrementally during training. This way the network is allowed to adjust to the quantization noise, minimizing the quantization error associated with the use of STE during QAT, which becomes especially apparent when quantizing to fewer than 8 bits. 

The idea of incremental quantization is taken even further in \cite{zhuang_towards_2018}, in which they apply QAT in several stages.  The authors introduce three methods of quantizing weights and activations:
\begin{enumerate}
    \item \textit{Two-Stage Quantization}: First, quantize the weights only, training them until they reach a sufficient accuracy. Then, do the same for the activations.
    \item \textit{Progressive Quantization}: Progressively decrease the bit widths for each QAT run. For instance, to quantize 32-bit floating point to 2-bit integer, first apply QAT to the 32-bit floating point model to quantize it to 16-bit integer. Then apply QAT to the 16-bit integer model to quantize it to 8-bit integer, continuing progressively by each lower power of 2 until the network is quantized to 2 bits.
    \item \textit{Guided Quantization}: Jointly train the full-precision model with the low-precision model based on knowledge distillation \cite{hinton_distilling_2015}, a (re)training process wherein both models learn from each other.
\end{enumerate}
These methods can be used independently or in combination. In fact, when they are all used together, 4-bit integer AlexNet and ResNet50 on ImageNet marginally outperform their 32-bit floating point counterparts. Even more, their 2-bit integer AlexNet on CIFAR-100 incurs less than 1\% accuracy loss.

\section{Binary and Ternary Quantization}\label{sec:binary-ternary}
In the previous section, we reviewed \cite{zhuang_towards_2018}, which showed results for quantizing NNs to 2-bit integers. When a network reaches such low bit widths, researchers realized that they could limit the representable values to take advantage of bit-wise operations, e.g., XNOR, which are extremely cheap to compute in hardware. NNs quantized in this manner fall into the category of binary and ternary quantization. Binary quantization involves 1 bit of precision, where the two representable values are typically $\{-1, +1\}$. Given these values, it is possible to compute nearly all of a NN's computations using addition, subtraction, logical bit shifts, and simple bit-wise operations. Such low precision often incurs high accuracy loss though, so researchers developed ternary networks, introducing a second bit of representation. With 2 bits of precision, it becomes possible to also represent 0, so a ternary network's values belong to $\{-1, 0, +1\}$, helping to recover some of the accuracy loss. Although it is possible to represent four values with ternary networks, this limited range allows for the use of bit-wise operations, which at the hardware-level is even lower-cost to execute than integer arithmetic. 

\subsection{Binary Quantization}
BinaryConnect \cite{courbariaux2015binaryconnect} brought the area of binary NNs to the forefront in 2015. The idea behind BinaryConnect is to constrain the weights to $\{-1, +1\}$ during the forward and backward training passes. \footnote{The weight updates are still performed using floating point gradients to allow Stochastic Gradient Descent to work properly and minimize accuracy loss.} The weights are stochastically binarized, in which weight $w$ is binarized to $w_b$ via

\begin{align}\label{eq:binaryconnect}
    w_b &= 
        \begin{cases}
            +1 & \text{with probability } p = \sigma(w) \\
            -1 & \text{with probability } 1 - p
        \end{cases}
\end{align}
where $\sigma$ is the ``hard sigmoid'' function:

\begin{align}
    \sigma(x) &= clip(\frac{x + 1}{2}, 0, 1) = max(0, min(1, \frac{x + 1}{2}))
\end{align}
Their experiments found stochastic binarization to be more effective than deterministic binarization, where $w_b$ is $+1$ if $w \geq 0$ and $-1$ otherwise. It is not immediately clear why binarization works, though the authors suggest that binarization acts as a regularizer by adding noise while maintaining the expected value of what the original weight would be at the end. Regularizers are typically used in NN training to reduce over-fitting and improve generalization. Although the goal of BinaryConnect is to speed up training using specialized hardware, it has the added benefit that the resulting binary NN has is more efficient during inference because a large share of the multiply-accumulate operations can be done via simple additions and subtractions; though, the remaining non-binarized parts of the network, such as the activations, still require floating point operations. 

\cite{hubara_binarized_2016} introduces BinaryNet, which binarizes activations in addition to the weights to $\{-1, +1\}$ according to Equation \ref{eq:binaryconnect}. To simplify training-time computation, they use a deterministic binarization for the weights because they found it to have negligible impact on accuracy. Given real weight $w$, we binarize it to $w_b$ via the following the deterministic binarization function:
\begin{align}\label{eq:sign}
    w_b &= Sign(w) = 
    \begin{cases}
        +1 & \text{if } w \geq 0 \\
        -1 & \text{otherwise}
    \end{cases}
\end{align}
With binarized weights and activations, the network can replace multiplications with XNOR and bit-counting operations in the multiply-accumulates as follows 
\begin{align}
    a_b' &\text{ += } popcount(XNOR(a_b, w_b))
\end{align}
where $a_b$ is the binarized input activation, $w_b$ is the binarized weight, and $a_b'$ is the output activation. The $popcount(\cdot)$ function counts the number of 1-bits. This equation is functionally equivalent to $a_b' += w_b \cdot a_b$, when the weights and activations are constrained to -1 and +1, while being significantly cheaper to compute than with multiplier units. In spite of these hardware-efficiency gains, they do not binarize the first layer. They argue that the first layer is often smaller than the other layers in the network, so paying this floating point cost is reasonable; however, as we have previously noted, this is not possible on hardware that lack floating point units, as is the case with many embedded systems.

\cite{leibe_xnor-net_2016} also applies the XNOR-popcount multiply-accumulate idea to their binary networks they call XNOR-Networks. They also introduce Binary Weight Networks (BWN), which are in contrast with XNOR-Nets in that their activations are not binarized, meaning they cannot use XNOR-popcount but use simple additions instead like BinaryConnect. BWN and XNOR-Nets differ from \cite{courbariaux2015binaryconnect, hubara_binarized_2016} because they use a different binarization function, which introduces a real-valued scaling factor. Both BWN and XNOR-Nets are binarized such that a real-valued weight $w$ is approximated as follows
\begin{align}
    w &\approx \alpha w_b
\end{align}
where $\alpha \in \mathbb{R}^+$ and $w_b \in \{-1, +1\}$, effectively binarizing the network to $\{-\alpha, +\alpha\}$. They compute $w_b$ using the simple $Sign(\cdot)$ function defined in Equation \ref{eq:sign} and $\alpha$ by taking the average of the absolute values of the weights in a given weight matrix. Since they use a real floating point scaling factor, they must multiply the scaling factor to the result the XNOR-popcount computations for each weight matrix. XNOR-Nets achieve better accuracy results than BWNs, BinaryConnect, and BinaryNets do because they use a scaling factor and a non-standard layer ordering.\footnote{They use BN-Activation-Convolution-Pool layer ordering as opposed to the standard Convolution-BN-Activation-Pool order.}

\subsection{Ternary Quantization}
Despite all of these inroads made in binary NNs, they still suffer from high accuracy loss (around 10-20\%) on challenging image classification tasks like ImageNet. As such, researchers have turned towards ternary NNs.

Ternary Weight Networks (TWNs) \cite{li_ternary_2016} build on BWNs and XNOR-Nets, as they also using a scaling factor, but this time using 2 bits of precision, limiting their weights to $\{-1, 0, +1\}$. To determine if a weight should be -1, 0, or +1, they set a threshold $\Delta$. They frame finding the threshold as an optimization problem that aims to minimize the Euclidean distance between the ternary weights and the original floating point weights, striking a balance between binary NNs and floating point NNs. The approximate the solution to the threshold optimization problem based on the assumption that the weights are normally distributed, yielding $\Delta \approx  0.7 \cdot \text{E}(|W|) \approx \frac{0.7}{n}\sum_{i + 1}^n |w_i|$, where $W$ is a weight matrix, and $n$ is the size of the matrix. Details of the optimization problem can be found in the paper. Based on this threshold, they convert a floating point weight $w$ to its ternary counterpart $w_t$ via
\begin{align}\label{eq:ternary}
    w_t &=
        \begin{cases}
            +1 & \text{if } w > \Delta \\
            0 & \text{if } |w| \leq \Delta \\
            -1 & \text{if } w < -\Delta
        \end{cases}
\end{align}
They compute the scaling factor $\alpha$ based on the threshold, in which $\alpha$ is the average of the absolute values of the weights greater than the threshold. The network thus uses ternary values $\{-\alpha, 0, +\alpha\}$. Based on this ternary quantization scheme, TWNs outperform binary NNs, incurring only 4\% accuracy loss with a TWN ResNet18 on ImageNet.

Rather than approximating the threshold for quantizing weights to ternary precision, Trained Ternary Quantization (TTQ) \cite{zhu_trained_2017} learns the ternary values and threshold for ternary assignment during training. Instead of choosing one scaling factor per weight matrix (as seen in BWN, XNOR-Nets, TWNs), TTQ is free to choose two different scaling factors; one for the lower bound and one for the upper bound. This means TTQ quantizes a NN to ternary values $\{-W^n_l, 0, +W^p_l\}$ for each layer $l$, in which $-W^n_l$ and $+W^p_l$ are floating point scaling factors that are learned during training. Since $-W^n_l \neq +W^p_l$, TTQ is an example of asymmetric quantization. The TTQ method involves
\begin{enumerate}
    \item normalizing the full-precision weights to $[-1, +1]$,
    \item quantizing weights to $\{-1, 0, +1\}$ (Equation \ref{eq:ternary}), and
    \item training the network.
\end{enumerate}
While TWN approximates the threshold for quantizing the weights, TTQ learns the threshold during retraining, in addition to $-W^n_l$ and $+W^p_l$. By using asymmetric $-W^n_l$ and $+W^p_l$ scaling factors, TTQ NNs are more flexible and have more model capacity. Using asymmetric quantization, however, comes at the cost of performing two separate multiplications per activation, rather than say a global element-wise multiplication. To account for this cost, it is possible to design custom hardware to have these asymmetric scaling factors pre-computed for the activations. TTQ is quite impressive because their TTQ ResNets outperform floating point ResNets on CIFAR-10 by $<$ 0.5\%.

\section{Mixed Precision Quantization}\label{sec:mixed}
The main issue with many of the previously discussed quantization schemes is that all the NN's parameters must be quantized to a single uniform precision. Although we saw some work in Section \ref{sec:int-fixed} choose different bit widths for weights, biases, and activations, the bitwidths were still uniform in each type of parameter. On the one hand, with integer/fixed point quantization, the precision of the weights, for instance, was held back by the most sensitive layer that needed the most bits when there may have been less sensitive layers that did not need as many. This reduces the achievable compression rate---we could have a smaller network with the same accuracy if we allocated fewer bits to the less sensitive layers. On the other hand, with binary and ternary quantization, the NNs were limited to 1 and 2 bits of precision, causing some NNs to suffer from high accuracy loss when they could have paid a few more bits to reduce accuracy loss.

Mixed precision attempts to address these problems by combining the previous quantization schemes together. Mixed precision takes the idea of using different bit widths for weights, biases, and activations even further, lowering the quantization bit width granularity to the layer-level. In effect, each layer has a tailored bit width and precision because mixed precision recognizes that some layers may benefit from more bits whereas others can afford to use fewer bits. Layers that are less sensitive to low-precision quantization are allocated fewer bits than the more sensitive layers. As a result, we meet integer/fixed point, binary, and ternary quantization in the middle, reducing accuracy loss at higher compression rates and more efficient inference.

The primary challenge of mixed precision is determining which bit widths are optimal for each layer because the search space is exponential in the number of layers. Brute-force searching this space is impractical, especially for deep networks. For example, quantizing ResNet50 to mixed precision bit settings where the possible bit widths are $\{1, 2, 4, 8\}$ has a search space of $4^{50} \approx 1.3 \times 10^{30}$ \cite{dong_hawq-v2_2019}. Hence, the principal challenge mixed precision research addresses is how to search this space efficiently or provide a principled framework that circumvents searching such a large space.

In \cite{lin_fixed_2016}, they use PTQ to quantize NNs to various fixed point schemes, in which they choose the optimal number of fractional bits that each fixed point precision has per layer. Recall from Section \ref{sec:background}.2 where we discuss how the \texttt{ap\_fixed}$<T,I>$ fixed point type, where $T$ is the total number of bits and $I$ is the number of integer bits. In \cite{lin_fixed_2016}, they are choosing the number of $T-I$ bits to use, which is the number of bits to allocate for the fractional part of the value, in effect tailoring the global fixed exponent that is used for each layer. To choose the optimal number of fractional bits, they frame finding the optimal bit widths as an optimization problem, in which they attempt to maximize what they define as the ``signal-to-quantization-noise-ratio'' (SQNR). After finding the best bit width, they compute the fractional bits needed. SQNR is based on the distance between quantized values and their original floating point values. They assume that more quantization noise (i.e., the further away the quantized values are from their original floating point values), the more the classification accuracy will degrade. After running a sufficient amount of calibration data through their network, they collect SQNR results to determine the bit width and compute the number of fractional bits needed. Their results show that mixed precision outperforms their uniform precision models.

\begin{table*}[h]
    \caption{\label{tab:schemes} A qualitative comparison of binary, ternary, integer/fixed point, and mixed precision quantization schemes}
    \centering
    \begin{tabular}{p{3cm} p{3cm} p{4cm} p{4cm}}
        \hline
        \textbf{Quantization Scheme} & \textbf{Accuracy Loss} & \textbf{Advantages} & \textbf{Disadvantages} \\
        \hline
        Binary & High & \textbf{Low cost.} All arithmetic done via binary operations. $32 \times$ size compression rate. & \textbf{High accuracy loss.} Binary networks often incur around 10\% accuracy reductions.  \\
        \hline
        Ternary & Low - Moderate & \textbf{High compression rate.} Multiplications done via binary operations or capped at two multiplications per activation if using asymmetric scaling factors. $16 \times$ compression rate. & \textbf{Floating point arithmetic}. For negligible accuracy loss, ternary networks use asymmetric floating point scaling factors, so they need to perform two floating point multiplications per activation. \\
        \hline
        Integer/Fixed Point & Low & \textbf{Integer arithmetic.} All arithmetic done via integer arithmetic, which is much cheaper than floating point arithmetic. & \textbf{Uniform precision.} For minimal accuracy loss, the networks are limited to the bit width of most sensitive layer, which is often 8 bits, so the compression rates are at most $4\times$. \\
        \hline
        Mixed Precision & Low & \textbf{Custom precision.} Quantization scheme for each layer or even row of weights is tailored to their precision sensitivity, reaping the benefits of binary, ternary, and integer quantization. & \textbf{Large search space.} The search space for which quantization scheme to use for each layer or weight row is exponential in the number of layers or weight rows, respectively. \\
        \hline
    \end{tabular}
\end{table*}

Hessian-Aware Quantization (HAWQ) \cite{dong_hawq_2019} also provides a systematic way during QAT to determine the precision of each layer's weights and activations while maintaining or improving current state-of-the-art quantization accuracy results. They use second-order information (the second derivative, or in this case specifically the second-order partial derivative), which for matrices is called the Hessian matrix (a matrix of the second derivatives), to determine how sensitive the weights and activations are. Based on this information, they determine the minimum bit width each layer needs to maintain overall network accuracy. The key observation is that layers with \textit{higher Hessian spectrum} (larger eigenvalues) have a more volatile loss. These layers are prone to more fluctuations in the loss when even a small amount of quantization noise is introduced (e.g., by rounding errors). Thus, they are more sensitive to quantization need a higher bit width. Layers with lower eigenvalues mean their loss is rather flat, even when larger amounts of quantization noise are introduced. With this in mind, these layers are less sensitive to quantization and can afford to have fewer bits. The idea is that a flat loss magnifies noise, e.g., quantization noise, significantly less than a region with sharper curvature in their loss. Based on this Hessian information, they manually select the bit widths for each layer. Another key insight from HAWQ is that the order in which layers are quantized is important and affects accuracy loss. They elect to quantize layers with higher Hessian values and a larger number of parameters, which means these layers are more sensitive to noise compared with the rest, and retrain them first before quantizing the remaining less sensitive layers. They argue that quantizing and retraining the less sensitive layers first is not very effective. The less sensitive layers adjust well to the introduction of quantization noise, so it is better to ``lock in'' the quantized values of the more sensitive layers first, allowing the less sensitive layers to recalibrate during this time. Even if this recalibration causes the parameters to stray further away from their original floating point values, their robustness allows them to still be quantized at this point with little effect on the network's overall accuracy. 

HAWQ sees two follow-ups in HAWQ-V2 \cite{dong_hawq-v2_2019} and HAWQ-V3 \cite{yao2021hawqv3}. The first follow-up HAWQ-V2 improved on HAWQ by using a better sensitivity metric and automatically selecting the bit widths for each layer. Instead of using the top Hessian eigenvalue as HAWQ does, HAWQ-V2 takes the average of all the Hessian eigenvalues of say a weight matrix to better capture how sensitive the layer is, rather than making decisions based on the layer's most sensitive parameter, i.e., its top eigenvalue. Based on this information, they take a Pareto frontier approach to automatically select bit width settings for each layer. Based on the average Hessian traces, they constrain the mixed precision search space and sort the candidate bit width settings based on their total second-order perturbation. This metric is defined as
\begin{align}\label{eq:second-order-pert}
    \Omega = \sum_{i = 1}^L \Omega_i = \sum_{i = 1}^L \overline{Tr}(H_i) \cdot ||Q(W_i) - W_i||_2^2
\end{align}
where $L$ is the number of layers, $\overline{Tr}(H_i)$ is the average Hessian trace, $||Q(W_i) - W_i||_2^2$ is the $L_2$ norm of the distance between quantized weights $Q(W_i)$ and floating point weights $W_i$. They argue that the bit width setting with minimum total second-order perturbation will generalize to the task at hand better, thus incurring low accuracy loss. This is not the optimal bit width setting, but it outperforms state-of-the-art manually selected bit width settings. 

The second HAWQ follow-up HAWQ-V3 \cite{yao2021hawqv3} makes significant improvements in that they eliminate all floating point operations in their Hessian-aware quantization scheme. In many quantization schemes that we have previously seen, including HAWQ and HAWQ-V2, floating point scaling factors were used. In HAWQ-V3, they propose to use \textit{dyadic numbers} as the scaling factors. Dyadic numbers are real numbers that can be represented as $b / 2^c$, where $b$ and $c$ are both integers. With dyadic scaling factors, the scaling factor multiplications and divisions can be done via integer multiplication and bit shifting, completely eliminating the need to support floating point numbers. Moreover, they further augment their Hessian-Aware quantization by making it more hardware-aware. They use Integer Linear Programming to find a mixed precision scheme that minimizes a NN's second-order perturbation (Equation (\ref{eq:second-order-pert})) subject to limits on the model size, number of binary operations, and latency. This makes the resulting models more practical for deployment on edge devices. 

AdaQuant \cite{hubara_accurate_2021} is a PTQ procedure that also uses ILP to determine their mixed precision quantization scheme, though they do not use the average Hessian trace in their optimization setup. With a small calibration dataset, they instead use the following objective to quantize each layer to its optimal precision:
\begin{align}
    (\hat\Delta_w, \hat\Delta_x, \hat V) &= argmin ||WX - Q_{\Delta_w}(W')Q_{\Delta_x}(X)||^2
\end{align}
In this equation, $\hat\Delta_w$ and $\hat\Delta_x$ are the step sizes for the weights and activations respectively, which determine their quantization scaling factors. $W$ is the given layer's weights, $X$ is the layer's input activations, and $Q(\cdot)$ is the quantization function. $W' = W + V$ where $V$ is a continuous variable to give the network some leeway during training, in which the quantized values need not be close to the original floating point values. As such, the quantized weights are defined as $W_q = Q_{\hat\Delta_w}(W + \hat V)$, giving the weights some space to account for quantization rounding errors. This equation can be run in parallel for each layer. To further correct the bias introduced by quantization, they run knowledge distillation using their calibration data. They also note that the common practice of fusing the BN layers with their predecessor weight layers (Equation (\ref{eq:norm})) \textit{before} applying PTQ is problematic, as seen in \cite{gong_highly_2018}. Before quantization, the BN parameters are reflecting the internal statistics of the \textit{floating point} model, not the quantized model. Therefore, they introduce ``Para-Normalization,'' a method to update the BN statistics according to the newly quantized model. They run a few forward passes of the calibration data through the quantized model, collecting new BN parameter statistics, and then re-fuse these new parameters into the weights, biases, and $\hat\Delta_w$ to adjust the weights' quantization scaling factor accordingly. 

While the previously discussed work \cite{hubara_accurate_2021,lin_fixed_2016, yao2021hawqv3, dong_hawq-v2_2019, dong_hawq_2019} takes a more systematic approach, others \cite{gong_mixed_2019, wang_haq_2019, uhlich_mixed_2020, yang_bsq_2021} leverage machine learning to address the challenge of mixed precision's large search space. 

\cite{wang_haq_2019, gong_mixed_2019} are more heavy-handed in their approaches. Hardware-Aware Quantization (HAQ) \cite{wang_haq_2019} uses reinforcement learning that takes hardware simulator results on latency and energy into account to satisfy the given resource constraints to find the optimal mixed precision bit width. \cite{gong_mixed_2019} searches for the optimal combination of NN architecture and quantization scheme by adding quantization as a search parameter during Neural Architecture Search (NAS). NAS involves automating the creation and search for new NN topological structures that outperform hand-designed ones. In this work, the authors claim that the optimal bit widths should be correlated with the architectures, so they should be search in conjunction to find more accurate and energy-efficient models. The search starts with MobileNetV2 \cite{sandler2018mobilenetv2} as a base model and $\{2, 4, 6, 8\}$ as the possible bit widths. They also use an energy simulator to obtain energy metrics for the NN models designed during NAS to guide their search. Their search algorithm finds that their mixed precision models achieve lower energy, lower latency, and lower accuracy loss than uniform precision quantization does. They also show that keeping NAS and quantization as separate processes yields models that are perform worse than their combined NN+Quantization search with respect to accuracy, model size, and energy efficiency. 

\begin{table*}[h]
    \caption{\label{tab:cifar10} Summary of various quantization methods on the CIFAR-10 dataset. In Precision ($w/a$), $w$ is the number of weight bits and $a$ is the number of activation bits. Bit-wise ops. = Bit-wise operations. MP = Mixed precision.}
    \centering
    \begin{tabular}{ccccccccc}
    Quantization & Method & Model & Acc. (\%) & Precision ($w/a$) & QAT/PTQ? & Bit-wise ops. & Int Arith. & Float Arith. \\
    \hline
    Int/Fixed Point & \cite{courbariaux_training_2015} & Maxout & 84.02 & 20/20 & QAT & & \checkmark & \\
    \hline
    \multirow{4}{*}{Binary \& Ternary} & BinaryConnect \cite{courbariaux2015binaryconnect} & CNN & 91.73 & 1/float32 & QAT & & & \checkmark \\
    & BNN \cite{hubara_binarized_2016} & CNN & 88.6 & 1/1 & QAT & \checkmark & & \checkmark \\
    & TWN \cite{li_ternary_2016} & VGG-7 & 92.56 & 2/float32 & QAT & \checkmark & & \checkmark \\
    & TTQ \cite{zhu_trained_2017} & ResNet56 & 93.56 & 2/float32 & QAT & \checkmark & & \checkmark \\
    \hline
    \multirow{2}{*}{Mixed Precision} & \cite{lin_fixed_2016} & AlexNet & 93.18 & 8/16 & QAT & & \checkmark & \\
    & HAWQ \cite{dong_hawq_2019} & ResNet20 & 92.22 & MP/4 & QAT & & \checkmark & \\
    \hline
    \end{tabular}
\end{table*}

\begin{table*}[h]
    \caption{\label{tab:imagenet} Summary of various quantization methods on the ImageNet dataset. In Precision ($w/a$), $w$ is the number of weight bits and $a$ is the number of activation bits. Bit. ops. = Bit-wise operations. MP = Mixed precision. * means this model was designed via NAS.}
    \centering
    \begin{tabular}{ccccccccc}
    Quantization & Method & Model & Acc. (\%) & Precision ($w/a$) & QAT/PTQ? & Bit. ops. & Int Arith. & Float Arith. \\
    \hline
    \multirow{14}{*}{Int/Fixed Point} & \cite{jacob_quantization_2018} & ResNet50 & 74.9 & 8/8 & QAT & & \checkmark & \\\cline{2-9}
    & Quant-Noise \cite{fan_training_2021} & EffNet-B3 & 79.8 & 8/8 & QAT & & \checkmark & \\\cline{2-9}
    & \multirow{2}{*}{\cite{zhuang_towards_2018}} & AlexNet & 58 & 4/4 & QAT & & \checkmark & \\
    & & ResNet50 & 75.7 & 4/4 & QAT & & \checkmark & \\\cline{2-9}
    & \multirow{2}{*}{\cite{gong_highly_2018}} & ResNet50 & 71.84 & 8/8 & PTQ & & \checkmark & \\
    & & Inception-V3 & 75.31 & 8/8 & PTQ & & \checkmark & \\\cline{2-9}
    & \multirow{4}{*}{LAPQ \cite{nahshan_loss_2020}} & ResNet18 & 68.8 & 8/4 & PTQ & & \checkmark & \\
    & & ResNet50 & 74.8 & 8/4 & PTQ & & \checkmark & \\
    & & ResNet101 & 73.6 & 8/4 & PTQ & & \checkmark & \\
    & & Inception-V3 & 75.1 & 8/4 & PTQ & & \checkmark & \\\cline{2-9}
    & \multirow{4}{*}{AdaRound \cite{nagel_up_2020}} & ResNet18 & 68.55 & 4/8 & PTQ & & \checkmark & \\
    & & ResNet50 & 75.01 & 4/8 & PTQ & & \checkmark & \\
    & & Inception-V3 & 75.76 & 4/8 & PTQ & & \checkmark & \\
    & & MobilenetV2 & 69.89 & 4/8 & PTQ & & \checkmark & \\\cline{2-9}
    \hline
    \multirow{5}{*}{Binary \& Ternary} & BWN \cite{leibe_xnor-net_2016} & ResNet18 & 60.8 & 1/float32 & QAT & & & \checkmark \\
    & XNOR-Net \cite{leibe_xnor-net_2016} & ResNet18 & 51.2 & 1/1 & QAT & \checkmark & & \\\cline{2-9}
    & TTN \cite{li_ternary_2016} & ResNet18 & 61.8 & 2/float32 & QAT & \checkmark & & \checkmark \\\cline{2-9}
    & \multirow{2}{*}{TTQ \cite{zhu_trained_2017}} & AlexNet & 57.5 & 2/float32 & QAT & \checkmark & & \checkmark \\
    & & ResNet18 & 66.6 & 2/float32 & QAT & \checkmark & & \checkmark \\
    \hline
    \multirow{17}{*}{Mixed Precision} & \multirow{3}{*}{AdaQuant \cite{hubara_accurate_2021}} & ResNet18 & 67.4 & MP & PTQ & & \checkmark & \\
    & & ResNet50 & 73.7 & MP & PTQ & & \checkmark & \\
    & & Inception-V3 & 72.6 & MP & PTQ & & \checkmark & \\\cline{2-9}
    & \cite{lin_fixed_2016} & AlexNet & 80 & MP & QAT & & \checkmark & \\\cline{2-9}
    & \multirow{2}{*}{HAWQ \cite{dong_hawq_2019}} & Inception-V3 & 75.52 & MP & QAT & & \checkmark & \checkmark \\
    & & ResNet50 & 75.48 & MP & QAT & & \checkmark & \checkmark \\\cline{2-9}
    & \multirow{2}{*}{HAWQ-V2 \cite{dong_hawq-v2_2019}} & Inception-V3 & 75.68 & MP & QAT & & \checkmark & \checkmark \\
    & & ResNet50 & 75.76 & MP & QAT & & \checkmark & \checkmark \\\cline{2-9}
    & \multirow{2}{*}{HAWQ-V3} & ResNet18 & 70.5 & MP & QAT & & \checkmark & \\
    & & ResNet50 & 75.95 & MP & QAT & & \checkmark & \\\cline{2-9}
    & \cite{gong_mixed_2019} & MobilenetV2* & 71.77 & MP & QAT & & \checkmark & \\\cline{2-9}
    & HAQ \cite{wang_haq_2019} & MobilenetV2 & 71.89 & MP & QAT & & \checkmark & \\\cline{2-9}
    & \multirow{2}{*}{\cite{uhlich_mixed_2020}} & MobilenetV2 & 70.59 & MP & QAT & & \checkmark & \\
    & & ResNet18 & 70.66 & MP & QAT & & \checkmark & \\\cline{2-9}
    & \multirow{2}{*}{BSQ \cite{yang_bsq_2021}} & ResNet50 & 75.29 & MP & QAT & & \checkmark & \\
    & & Inception-V3 & 76.6 & MP & QAT & & \checkmark & \\
    \hline
    \multirow{2}{*}{Mixed Scheme} & \multirow{2}{*}{\cite{chang_mix_2021}} & ResNet18 & 70.27 & 4/4 & QAT & \checkmark & \checkmark & \\
    & & MobilenetV2 & 71.31 & 4/4 & QAT & \checkmark & \checkmark & \\
    \hline
    \end{tabular}
\end{table*}

\cite{uhlich_mixed_2020, yang_bsq_2021} take a more traditional QAT approach when finding the best mixed precision schemes by learning the best mixed precision quantization parameters during QAT. \cite{uhlich_mixed_2020} claims that learning the quantization function's parameters is possible if a good parameterization is chosen during training. They ascertain that a good parameterization to learn during training include the step size and dynamic range of the the quantizer, whereas learning the bit width itself performed worse. Instead, the bit width is inferred from learned step size and dynamic range. This method learns the step size and dynamic range for each layer, leading to a mixed precision quantization. The authors also note that starting with a pre-trained floating point model outperforms starting from scratch with a random weight initialization. Bit-level Sparsity Quantization (BSQ) \cite{yang_bsq_2021} also uses a more standard QAT framework; however, the lower the granularity of quantizing at the layer level to the bit level. In their work, the authors treat each bit used to represent each weight as an independent variable, forcing some bits to 0 to induce sparsity and lower bit widths. BSQ induces negligible accuracy loss while achieving higher compression rates compared to previous mixed precision quantization methods, such as HAWQ \cite{dong_hawq_2019}, though they keep the precision of the first and last layers fixed at 8 bits.

\section{Future Directions}\label{sec:future}
Plenty of work has been done in quantization, as we have surveyed in this paper. As seen in Table \ref{tab:schemes}, the quantization schemes we have discussed each have their advantages and disadvantages. Additionally, Tables \ref{tab:cifar10} and \ref{tab:imagenet} present overviews of the accuracy and arithmetic the hardware is required to support. Based on this information, researchers can decide which quantization schemes best suit their deployment requirements. 

Nevertheless, there is plenty of opportunity for improvement to make NNs more easily deployed to hardware. Not a lot of work has focused on fixed point quantization and finding the optimal number of fractional bits to use for each layer \cite{lin_fixed_2016}. This might be because the fixed point data type is mainly used in the embedded systems space and few other fields of computer science. This is one sub-field of mixed precision quantization that could be further studied, especially considering executing fixed point data types in hardware is on par in terms of hardware-efficiency with executing integers. The main challenge is that the search space for how many fractional bits to use is large and also exponential in number of layers if we want to tailor it to each network. This challenge is the same challenge of mixed precision quantization, so it would be fruitful to apply methods of searching the mixed precision space to searching the fixed point precision space.

Additionally, little work has been done on combining the approaches together \cite{gholami_survey_2021}. While mixed precision does assign different bit widths for each layer, they are usually still all of the same data type. With custom hardware, it is possible to group layers together based on the best quantization scheme for them for more efficient hardware deployment. The key challenge here is to balance the complexity of the hybridized quantization schemes with the complexity it would take to implement them in hardware. It would be less than ideal to formulate a highly optimized hybrid quantization scheme whose implementation overhead outweighs the potential efficiency gained in hardware. \cite{chang_mix_2021} has shown results for combining their novel hardware-friendly quantization scheme called \textit{sum-of-power-of-2} with fixed point quantization---both these schemes are efficient to implement using FPGA Digital Signal Processing units (DSPs). Since they claim to be the first paper to combine quantization schemes, they dub this process \textit{mixed scheme quantization}. The optimal combination is learned during training. Thus, \cite{chang_mix_2021} shows promise for this research direction. 

In addition to combining quantization schemes together, there is also ample opportunity to explore combining quantization with other NN compression techniques, such as NAS, knowledge distillation, and pruning. Deep Compression \cite{he_deep_2015}, Quantization-Aware Pruning \cite{tran2021ps}, and TTQ \cite{zhu_trained_2017} have shown results in combining fixed point and ternary quantization with pruning. Distillation-assisted quantization is also an emerging field \cite{mishra_apprentice_2017, yao2021hawqv3}. But, there are more quantization schemes that could benefit from combining with other compression techniques. Moreover, there is little work on what the optimal combination of compression techniques is. 

NN quantization is a well-studied topic with many degrees of freedom with plenty of future directions for the field. The greatest advantage of quantization is the efficiency manifested in hardware. With more models that can be deployed on hardware and edge devices, the greater the impact of NNs will be and the more humanity will reap its benefits.

\bibliographystyle{ACM-Reference-Format}
\bibliography{refs}

\end{document}